\documentclass[sigconf,screen]{acmart}

\pdfoutput=1

\usepackage{comment}
\usepackage{multirow, makecell}
\usepackage{soul}
\usepackage{pifont}
\usepackage{enumitem}

\setcopyright{acmcopyright}
\copyrightyear{2022}
\acmYear{2022}
\acmDOI{XXXXXXX.XXXXXXX}

\acmBooktitle{Woodstock '18: ACM Symposium on Neural Gaze Detection,
 June 03--05, 2018, Woodstock, NY} 

\begin{document}

\settopmatter{printacmref=false}
\setcopyright{none}
\renewcommand\footnotetextcopyrightpermission[1]{}
\pagestyle{plain}

\title{In Search of Verifiability: Explanations Rarely Enable Complementary Performance in AI-Advised Decision Making}

\author{Raymond Fok}
\email{rayfok@cs.washington.edu}
\affiliation{
  \institution{University of Washington}
  \city{Seattle}
  \state{WA}
  \country{USA}
}

\author{Daniel S. Weld}
\email{danw@allenai.org}
\affiliation{
  \institution{Allen Institute for AI \&\ \\ University of Washington}
  \city{Seattle}
  \state{WA}
  \country{USA}
}

\renewcommand{\shortauthors}{}

\newcommand{\needcite}[1]{\textcolor{purple}{[XX CITE XX]}}

\newcommand\ray[1]{\textcolor{blue}{[Ray]: #1}}
\newcommand\dan[1]{\textcolor{red}{[Dan]: #1}}

\newcommand\edit[1]{{{#1}}}

\newcommand{\Bug}
    {\mbox{\rule{2mm}{2mm}}}
\newcommand{\DSW}[1]
    {\bug \footnote{\textcolor{red}{\textit{DSW: #1}}}}
    
\definecolor{dukeblue}{rgb}{0.0, 0.0, 0.61}
\definecolor{darkspringgreen}{rgb}{0.09, 0.45, 0.27}
\definecolor{fireenginered}{rgb}{0.81, 0.09, 0.13}

\newcommand{\cmark}{\textcolor{darkspringgreen}{\ding{51}}}
\newcommand{\xmark}{\textcolor{fireenginered}{\ding{55}}}
\newcommand{\asterisk}{\textcolor{dukeblue}{$\star$}}

\begin{abstract}
The current literature on AI-advised decision making---involving explainable AI systems advising human decision makers---presents a series of inconclusive and confounding results. To synthesize these findings, we propose a simple theory that elucidates the frequent failure of AI explanations to engender appropriate reliance and complementary decision making performance. In contrast to other common desiderata, e.g., interpretability or spelling out the AI's reasoning process, we argue explanations are only useful to the extent that they \textit{allow a human decision maker to verify the correctness of the AI's prediction}. Prior studies find in many decision making contexts AI explanations \textit{do not} facilitate such verification. Moreover, most tasks fundamentally do not allow easy verification, regardless of explanation method, limiting the potential benefit of any type of explanation. We also compare the objective of complementary performance with that of appropriate reliance, decomposing the latter into the notions of outcome-graded and strategy-graded reliance.
\end{abstract}




\maketitle

\section{Introduction}

\begin{figure}[t]
    \centering
    \vspace{1em}
    \includegraphics[width=0.47\textwidth]{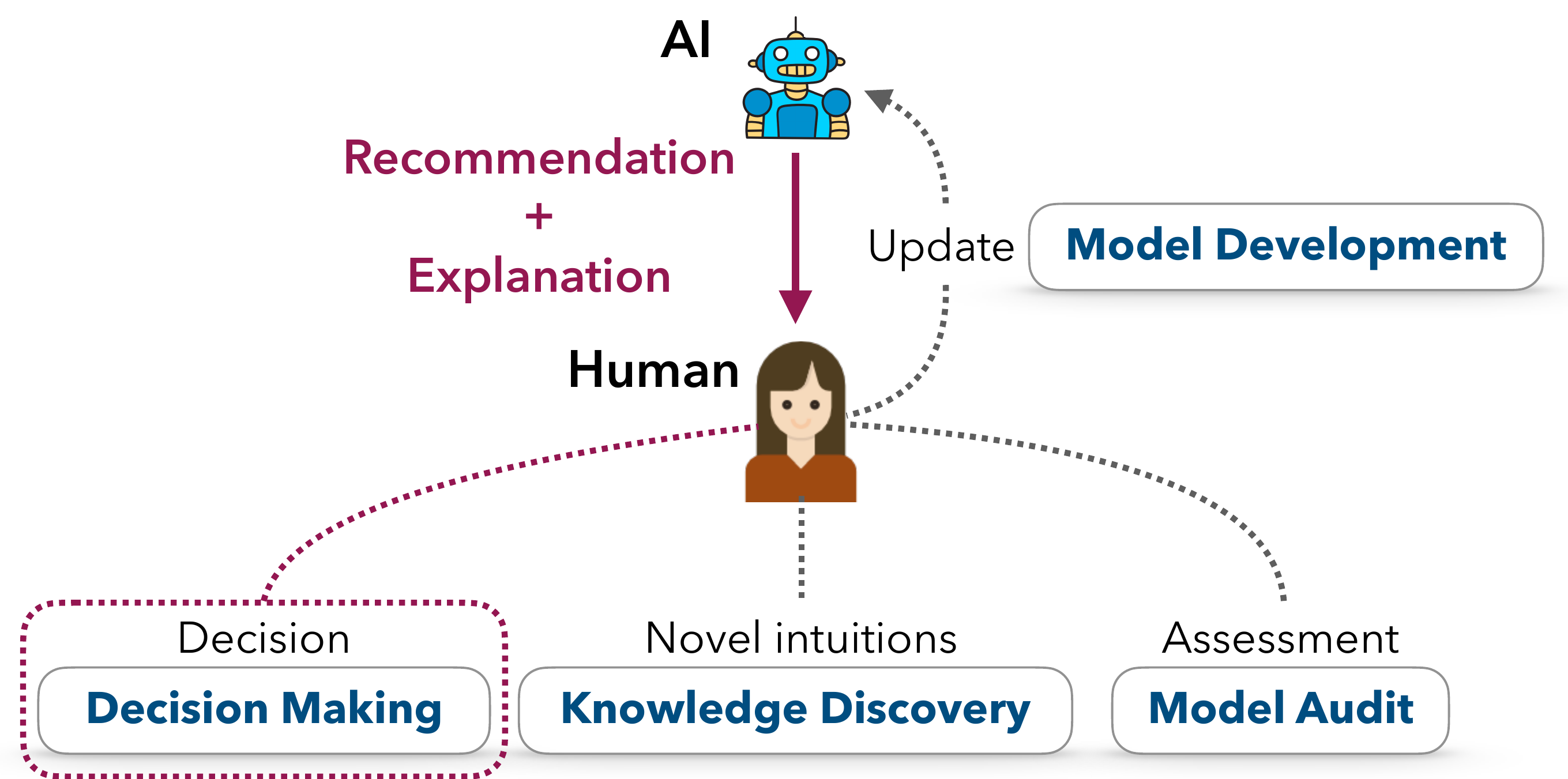}
    \caption{Researchers suggest that AI explanations could aid numerous human-AI processes, including decision making, model development, knowledge discovery, and model audit. In this paper, we focus solely on understanding whether explanations are helpful in the context of AI-advised decision making. We claim AI explanations cannot foster appropriate reliance and engender complementary performance in decision making, except in the rare instances in which they efficiently verify the AI's recommendation.}
    \Description{}
    \label{fig:xai_contexts}
\end{figure}
\renewcommand\theadfont{\normalfont\bfseries}

\begingroup

\begin{table*}[t]
    \small
    \centering
    \begin{tabular}{l l c c c}
    \toprule
    \makecell[l]{\textbf{Paper}} &
    \makecell[l]{\textbf{Decision Making Task}} &
    \makecell{\textbf{Explanations}} &
    \makecell{\textbf{Enables}\\\textbf{Verification}} &
    \makecell{\textbf{Complementary}\\\textbf{Performance}} \\
    \hline
    \citet{biran_human-centric_2017} & Stock price prediction & Rationale & \xmark & \xmark \\
    \citet{green_principles_2019} & Pretrial detention & Feature importance & \xmark & \xmark \\
    \citet{green_principles_2019} & Financial lending & Feature importance & \xmark & \xmark \\
    \citet{weerts_human-grounded_2019} & Income prediction & Feature importance & \xmark & \xmark \\
    \citet{lai_on_2019} & Deceptive review & Feature importance (text highlights) & \xmark & \xmark \\
    \citet{lai_why_2020} & Deceptive review & Feature importance (text highlights) & \xmark & \xmark \\
    \citet{bucinca_proxy_2020} & Food fat content prediction & Feature importance & \xmark & \xmark \\
    \citet{carton_feature-based_2020} & Toxicity detection & Feature importance (text highlights) & \xmark & \xmark \\
    \citet{zhang_effect_2020} & Income prediction & Feature importance & \xmark & \xmark \\
    \citet{liu_understanding_2021} & Recidivism prediction & Feature importance & \xmark & \xmark \\
    \citet{alufaisan_does_2021} & Income prediction & Feature importance & \xmark & \xmark \\
    \citet{alufaisan_does_2021} & Recidivism prediction & Feature importance & \xmark & \xmark \\
    \citet{wang_are_2021} & Recidivism prediction & Feature importance & \xmark & \xmark \\
    \citet{wang_are_2021} & Forest cover prediction & Feature importance & \xmark & \xmark \\
    \citet{nourani_anchoring_2021} & Kitchen policy verification & Feature importance & \xmark & \xmark \\
    \citet{poursabzi-sangdeh_manipulating_2021} & Apartment price prediction & Feature importance & \xmark & \xmark \\
    \citet{jesus_how_2021} & Fraud detection & Feature importance (text highlights) & \xmark & \xmark \\
    \citet{nguyen_effectiveness_2021} & Image classification & Attribution maps & \xmark & \xmark \\
    \citet{kim_hive_2022} & Image classification & Attribution maps \& Prototype-based & \xmark & \xmark \\
    \citet{parrish_singleturn_2022} & Long document MCQA & Rationale & \xmark & \xmark \\
    \citet{sivaraman_ignore_2023} & Sepsis treatment & Feature importance & \xmark & \xmark \\

    \hline
    \citet{bansal_doesthewhole_2021} & Sentiment classification & Feature importance (text highlights) & \xmark & \asterisk \\
    \citet{{bansal_doesthewhole_2021}} & Logical reasoning & Rationale & \xmark & \asterisk \\
    \citet{taesiri_visual_2022} & Image classification & Nearest neighbors & \xmark & \asterisk \\
    
    \hline
    \citet{feng_what_2019} & Quizbowl trivia QA & Rationale & \cmark & \cmark \\
    \citet{gonzalez_human_2020} & Wikipedia ODQA & Rationale & \cmark & \cmark \\
    \citet{vasconcelos_when_2023} & Maze solving & Path highlights & \cmark & \cmark \\
    \citet{lee_evaluating_2022} & Multitask MCQA & Rationale (interactive) & \cmark & \cmark \\
    \citet{bowman_measuring_2022} & Long form MCQA & Rationale (interactive) & \cmark & \cmark \\
    \bottomrule
  \end{tabular}
  \caption{Recent studies investigating the effect of AI explanations on task performance in AI-advised decision making. Common explanation types include descriptions of feature importance (over both tabular and textual features) and natural language rationales. Most of these explanations describe the AI's decision-making {\em process} rather than help a human decision maker \textit{verify} the AI's recommendation. But only the latter engender {\em complementary performance} (\cmark), where team performance exceeds that of the human or AI alone.
  Sometimes explanations which do not enable verification may appear to yield complementary performance (\asterisk); however, in these cases, it is unclear whether explanations actually offer utility as providing only AI recommendations and confidence scores (without the explanations) also resulted in complementary performance.}
  \label{tab:xai_studies}
\end{table*}

\endgroup

Recent years have seen an explosion of work on explainable AI (XAI), but there have been mixed results on whether explanations actually help humans who are making decisions with  AI support. In this decision making context, the role of explanations is to foster appropriate reliance by helping the human understand whether or not the AI’s advice should be trusted. Appropriate reliance is desired in order to achieve complementary performance, where the human-AI team performs better than either the human or AI alone~\cite{bansal_doesthewhole_2021}. But here we see a confusing montage of results: not only do most papers find explanations don’t induce complementary performance more than baseline methods, such as displaying AI accuracy or confidence, but these papers suggest explanations can in fact increase over-reliance, where the human trusts the AI even when it errs. The inconclusive nature of these results raises a huge question for the field of XAI: when are explanations useful?

We focus solely on the process of AI-advised decision making, defined as the following: given an instance of a decision making task, an AI makes a recommendation, and drawing on features of the task, the AI’s recommendation, and possibly an explanation for the AI’s recommendation, a human decision maker arrives at a final decision (Figure~\ref{fig:xai_contexts}). There are many other possible uses for AI explanations~\cite{tan_on_2022, kim_help_2023}, including model debugging and auditing, e.g., to help the human understand whether the AI's reasoning will generalize, but our arguments pertain only to decision making.

In this paper, we present a perspective we believe explains the seemingly mixed empirical results found throughout the XAI literature. Furthermore, our proposal is consistent with the way human groups reach consensus on ``intellective'' tasks~\cite{laughlin_1986_demonstrability}. \textbf{We argue explanations provided by an AI model are helpful in decision making (engender complementary performance~\cite{bansal_doesthewhole_2021}) to the extent they allow a decision maker to verify the AI’s recommendation.} While this theory may appear self-evident,
\begin{enumerate}
    \item Most work on XAI has focused instead on creating inherently interpretable models or generating faithful post-hoc explanations of the AI's reasoning process.
    \item Most of these explanations do not support such verification.
\end{enumerate}
Explanations which faithfully expose the AI's reasoning process may well be useful for debugging the AI or predicting its ability to generalize, but it does not seem to help human decision makers make judgements on individual task instances. Indeed, most human-subject studies have shown that explanations fail to produce complementary performance in decision making; the sole exceptions are explanations that support answer verification (Table~\ref{tab:xai_studies}).

The rest of this paper is structured as follows. The next section surveys the conflicting results from prior studies on XAI utility. Section~\ref{s:xadm} details the decision making context, which is our focus in this paper. Section~\ref{s:verify} presents our core argument --- explanations must facilitate verification in order to engender complementary performance. Section~\ref{s:ar} discusses the concept of appropriate reliance, arguing this term has become overloaded, leading to confusion; we propose two alternative terms: \textit{outcome-graded reliance} and \textit{strategy-graded reliance} to tease these apart. Section~\ref{s:discuss} places our conjecture in a broader context, and Section~\ref{s:conclude} concludes.
\section{Background} \label{s:background}

\begin{figure*}[t]
    \centering
    \includegraphics[width=\textwidth]{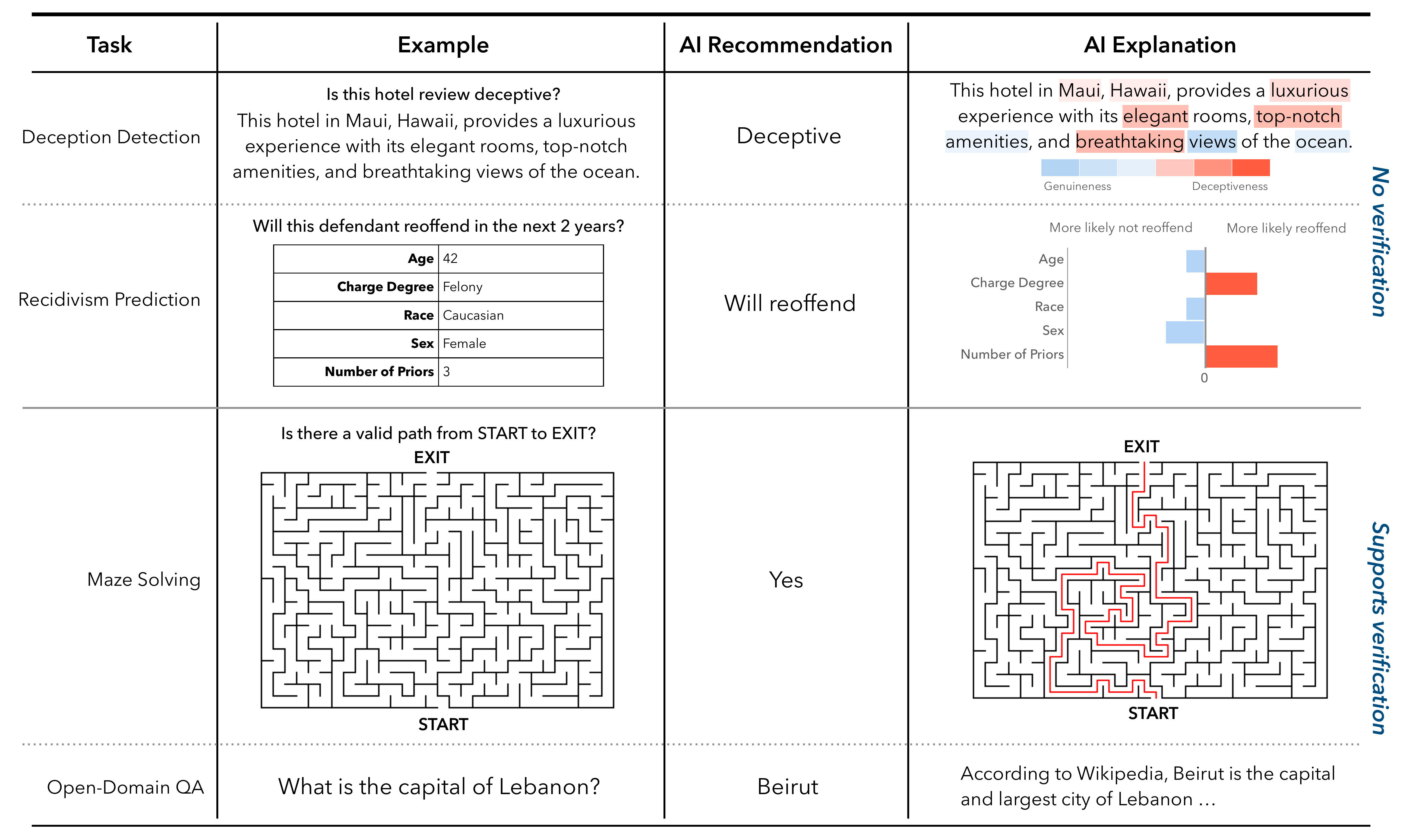}
    \caption{Examples of AI-advised decision making tasks. The top two rows present prototypical tasks found in the XAI literature: deceptive review detection, with textual features, and recidivism prediction, with tabular features. Both tasks often utilize feature importance explanations, which studies suggest \textit{rarely} engender complementary performance or foster appropriate reliance. The bottom two rows show examples of a maze solving and open-domain question answering task, for which studies have found AI explanations \textit{can} engender complementary performance.}
    \Description{}
    \label{fig:aiadm_examples}
\end{figure*}

Providing interpretability for AI models, for instance through explanations, has been one way researchers have attempted to facilitate more informed and accurate decision making~\cite{bilgic:iui-bp05}. One set of studies have found AI explanations can improve human-AI performance over human decision making alone and over human-AI teams with AI recommendations without explanations~\cite{eiband_impact_2019, lai_why_2020, bucinca_proxy_2020, bucinca_trustorthink_2021, biran_human-centric_2017, horne_rating_2019, lai_on_2019}. However, none of these studies found explanations could improve human-AI performance beyond the original capabilities of the AI model. The studies also involved collaborations in which the AI model significantly outperformed the human decision maker. It is therefore unclear whether explanations truly improved the human-AI decision making process, or if explanations convinced humans to blindly trust the AI's recommendations. In this type of scenario --- where the AI performs at a superhuman level --- should we simply allow the AI to operate independently? It is arguably naive to believe the human is performing any meaningful oversight, when the effect of explanations is inducing blind trust.

In response, researchers have investigated how explanations might support decision makers in developing \textit{appropriate reliance} on AI assistance, rather than improving performance through blind trust. While appropriate reliance is not well-defined in the literature (as explained in \S\ref{s:ar}), one common definition states that appropriate reliance is achieved when decision makers agree with the AI when its recommendation is correct, and disagree with the AI when its recommendation is incorrect. Unfortunately, achieving this notion of appropriate reliance through explanations has been elusive~\cite{bansal_doesthewhole_2021, chandrasekaran_explanations_2018, kiani_impact_2020, alqaraawi_evaluating_2020, jacobs_how_2021, carton_feature-based_2020, liu_understanding_2021, nourani_anchoring_2021, poursabzi-sangdeh_manipulating_2021, weerts_human-grounded_2019, zhang_effect_2020, jesus_how_2021, parrish_singleturn_2022, green_principles_2019, sivaraman_ignore_2023}. Fundamentally, it can be hard for people to know how much to trust recommendations~\cite{ghai_explainable_2021, jacovi_aligning_2021, sivaraman_ignore_2023, bucinca_proxy_2020}. And though AI explanations were hypothesized to make the AI's decision making process more interpretable and in turn support appropriate reliance, most experiments have found little evidence for such support. Explanations have not only often failed to foster appropriate reliance, many studies have found explanations can actually negatively impact appropriate reliance~\cite{bucinca_trustorthink_2021, bucinca_proxy_2020, gajos_do_2022, sivaraman_ignore_2023, kim_assisting_2023}.

For example, the mere presence of explanations can increase a decision maker's trust in the AI's recommendations~\cite{bucinca_trustorthink_2021, designing_wang_2019, yu_do_2019, zhang_effect_2020, kim_hive_2022} (even irrespective of explanation quality~\cite{eiband_impact_2019}), revealing fallacies in human decision-making --- most notably, confirmation bias.  If shown to human decision makers before they have a chance to formulate a decision themselves, explanations can further introduce an anchoring bias~\cite{ghai_explainable_2021, nourani_anchoring_2021, bansal_doesthewhole_2021, bowman_measuring_2022, bucinca_trustorthink_2021, designing_wang_2019}. Explanations may be effective in increasing decision makers' perceptions of an AI's usefulness and decision making confidence~\cite{sivaraman_ignore_2023, wang_are_2021, tschandl_humanComputer_2020, alam_examining_2021}, but these characteristics do not necessarily translate to improved decision making performance.

Another set of studies have found explanations fail to improve decision making performance~\cite{green_principles_2019, weerts_human-grounded_2019, zhang_effect_2020, alufaisan_does_2021, bansal_doesthewhole_2021, carton_feature-based_2020, liu_understanding_2021, poursabzi-sangdeh_manipulating_2021, jesus_how_2021, parrish_singleturn_2022, kim_assisting_2023}. One hypothesis for these negative results suggests decision makers do not actually cognitively engage with AI explanations~\cite{chromik_i_2021, eiband_impact_2019, ghai_explainable_2021, bunt_are_2012}, resulting in diminished utility of an explanation regardless of its type or quality. Another hypothesis for these negative results lies in the potentially low fidelity of post-hoc explanations for black-box models. Some argue these post-hoc explanations can provide inaccurate representations of the original model, and as a result, decision makers are burdened with determining the trustworthiness of not only the AI's recommendation but also its explanation~\cite{rudin_stop_2019, ghassemi_false_2021, mittelstadt_explaining_2019}. However, this supposed interpretability gap between intrinsically interpretable models and post-hoc interpretability has shown no significant impact on decision making performance~\cite{poursabzi-sangdeh_manipulating_2021, bell_notthatsimple_2022}.

A third set of studies suggests explanations can sometimes yield \textit{complementary performance}~\cite{bansal_doesthewhole_2021}, defined as when the decision making performance of the human-AI team exceeds the capabilities of either the human or AI alone~\cite{feng_what_2019, vasconcelos_when_2023, gonzalez_human_2020, lee_evaluating_2022, bowman_measuring_2022}. Unlike the first set of studies where overall human-AI performance lagged behind AI performance, these studies found explanations influenced decision makers into appropriately relying on the AI recommendations.

All in all, the current literature on XAI-advised decision making presents a smattering of intriguing empirical results. Some studies have found explanations \textit{cannot} foster appropriate reliance and fail to engender complementary performance, while others have found explanations \textit{can} achieve complementary performance. Rather than viewing these results as conflicting, we aim to provide a unifying theory that explains under what circumstances these explanations can in fact improve decision making performance, if at all.
\section{AI-Advised Decision Making}
\label{s:xadm}

We restrict our analysis of AI explanation utility to the established paradigm of \textit{AI-advised decision making}~\cite{bansal_updates_2019}, defined as follows:
\begin{enumerate}
    \item Given an individual instance of a task, an AI makes a recommendation, and possibly provides an explanation.
    \item A human decision maker makes the final decision, drawing on features of the task, an AI’s recommendation, and (if available) its explanation (Figure~\ref{fig:aiadm_examples}).
\end{enumerate}
Moreover, we are concerned with the effect of explanations on \textit{decision making performance} via objective measures of efficacy such as accuracy, error rate, or speed of decision-making~\cite{bansal_beyond_2019}.

Beyond simply improving the performance of a human-AI team, we are interested in \textit{complementary performance}, where the human-AI system performs better (for example, its accuracy is higher or decisions are made more cheaply) than either the human or AI separately. While complementarity is not necessary for AI assistance to be deemed ``useful,'' for many researchers it is the intuitive motivation for coupling a human and AI together. Furthermore, if the consequence of using a human-AI team \textit{lowers} performance (for whatever metric is deemed most important), we should have clear understanding of this effect.

Given the focus on complementary performance, we do not attempt to characterize the utility (or lack thereof) of explanations for other possible objectives in AI-advised decision making. These include task-centered metrics such as subjective assessments of efficacy (e.g., self-rated accuracy, decision confidence) or satisfaction (e.g., cognitive load, helpfulness), and AI-centered metrics such as trust and reliance on the AI, perceptions of AI fairness, and understanding of the AI.

Explanations are often characterized based on the scope of information they convey about an AI model. For decision making, the existing literature largely studies the influence of \textit{local explanations}, which provide information about individual AI recommendations. These include visualizations of model uncertainty, feature importance, rule-based explanations, example-based explanations, counterfactual explanations, or natural language rationale explanations. In contrast, \textit{global explanations} provide holistic insight into the entirety of an AI model, for instance by visualizing or detailing a model's complete architecture. Faithfulness is another dimension used to characterize explanations. A faithful explanation is one that accurately represents the reasoning process behind a model's prediction. Some intrinsically interpretable models, such as GA$^2$Ms~\cite{caruana_intelligible_2015}, are favored because it is (relatively) easy for a human to understand the model's behavior. Other explanations are generated by post-hoc analysis, e.g. LIME~\cite{ribeiro_whytrust_2016} or SHAP~\cite{lundberg_explainable_2018}. While these methods produce local explanations for otherwise inscrutable models, they are by definition approximations of a model's true reasoning process, and hence raise concerns of faithfulness.

In this paper, we present a theory of verification agnostic to these particular characterizations of explanations (local vs. global, intrinsically interpretable vs approximate post-hoc). We argue explanation faithfulness is largely irrelevant; the main issue that affects decision making performance is whether the explanation helps the human verify the proposed solution.
\begin{figure}[t]
    \centering
    \includegraphics[width=0.46\textwidth]{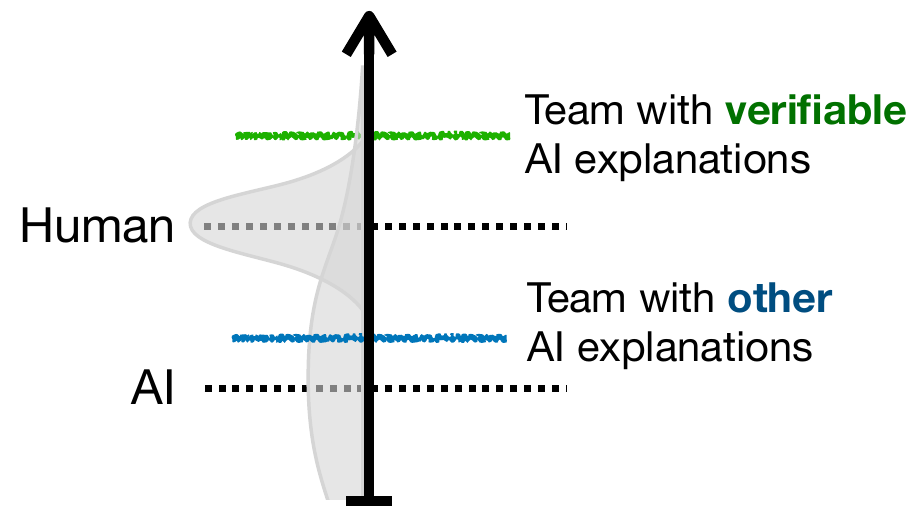}
    \caption{\edit{In decision making contexts where the AI outperforms the human, improved team performance can often be achieved by providing an AI explanation, as explanations tend to cause humans to trust the AI {\em regardless} of its correctness. Unfortunately, when the AI errs more than the human, these explanations can reduce performance due to over-reliance. However, explanations that support quick verification may still produce complementary performance by allowing the human to recognize situations where they have made a mistake---even if the AI is less accurate {\em on average} than the human.}}
    \Description{}
    \label{fig:verifiable_expl}
\end{figure}

\section{Verifiability} \label{s:verify}

\begin{figure*}[t]
    \centering
    \includegraphics[width=\textwidth]{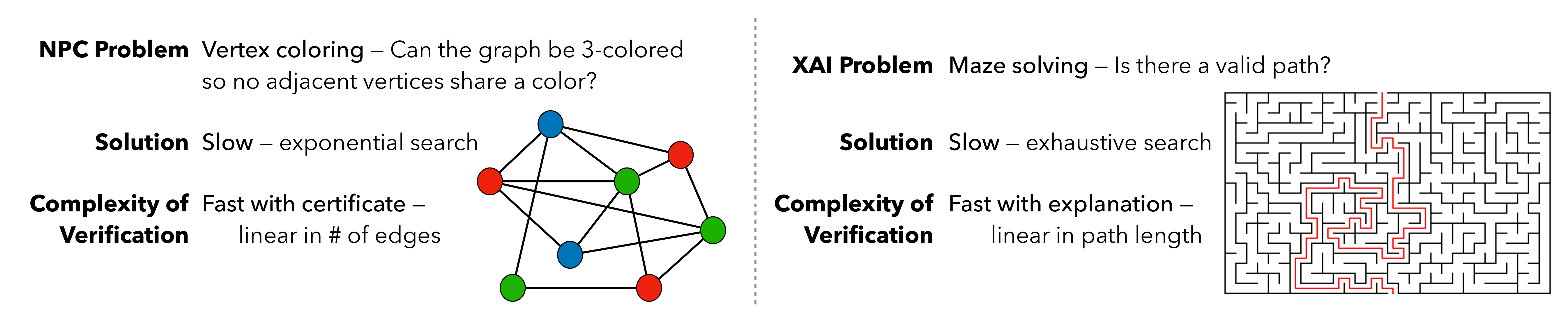}
    \caption{Tasks in which AI explanations have been shown to provide verification of an AI recommendation (e.g., maze solving) resemble NP-complete decision problems. Complementary performance can be achieved in these cases because the AI does the time-consuming problem solving and the human can quickly verify the correctness of a solution. Of course, this verification process is task specific. Sometimes fast verification is only possible for some answer types; for example, an illegal maze path does not preclude the existence of a legal path. On average, however, these explanations speed human verification~\protect{\cite{vasconcelos_when_2023}}.}
    \Description{}
    \label{fig:npcomplete_examples}
\end{figure*}

\textbf{We argue that explanations can enable complementary performance in AI-advised decision making to the extent they allow a decision maker to verify an AI’s recommendation.} Here, we refer to \textit{verification} of a candidate answer as the process of determining its correctness. In the psychology literature, this notion is also called \textit{demonstrability} and has been associated with the ability of one individual \edit{in a group} to convince other \edit{group} members to switch to a correct answer on `intellective' problems such as math puzzles~\cite{laughlin_1986_demonstrability}.

Note that some problems may not have a verifiable answer, and many AI explanations fundamentally cannot satisfy this desideratum and thus do not effectively support complementary decision making. For instance, one of the most common types of explanations is local feature importance, e.g., model coefficients of interpretable models or approximate post-hoc explanations. These types of explanations which describe a decision-making process can either align with human intuitions, potentially resulting in confirmation bias, or contradict human intuitions, in which case a decision maker may often find it more convenient to explain away any differences. In the best case, the AI's explanation reveals a salient misalignment between the AI's decision making process and its expected behavior, allowing a decision maker to disregard the AI's recommendation. While feature importance explanations may provide some indication of how much each feature influenced the AI's decision, they typically do not allow a decision maker to verify the AI's recommendation.

\subsection{Useful Explanations Allow Verification}
To see the importance of verification, consider the task of AI-advised maze solving (similar to the study in~\cite{vasconcelos_when_2023})---the goal is to determine whether a valid path exists between a specified entrance and exit in the maze. The AI recommends a binary decision (i.e., yes or no), and explains its decision with a highlighted path through the maze to the exit. Without the explanation, a human cannot verify the AI's recommendation short of solving the maze themselves. Thus the human must adopt a policy of blind trust or disregard the AI's advice---neither of which produce complementary performance. Given the path, however, the decision maker is able to easily verify the AI's recommendation, filtering mistakes, and only expending effort to solve mazes where the AI errs.
\edit{As shown in Figure~\ref{fig:verifiable_expl}, this can yield complementary performance even when the AI is overall less accurate than the human.}

Verification given AI explanations in decision making closely resembles verification of \textit{certificates} or solutions within canonical NP-complete computational decision problems (Figure~\ref{fig:npcomplete_examples}). While finding a solution to an NP-complete problem likely takes exponential time, verifying the certificate can be done with polynomial computation. In both contexts, while finding a solution is challenging (e.g., requiring an exhaustive search), checking the correctness of an explanation enables efficient verification of the answer.

Unfortunately, our maze example is unusual --- explanations in the tasks most commonly considered in XAI studies (e.g., recidivism prediction, medical applications, sentiment analysis, or deceptive review detection) do not enable verification. Explanations for these tasks typically include providing model coefficients or key features, which serve to elucidate the model's decision making \textit{process} rather than enable verification of a proposed decision.

For example, for a sentiment analysis task an AI model may explain itself by highlighting spans of text which were most influential to its decision (Figure~\ref{fig:compare} (right)). This may appear reassuring, but can lead to a false sense of trust when the AI makes mistakes. Indeed, user studies have shown that these explanations aren't any more helpful than seeing the AI's internal confidence~\cite{bansal_doesthewhole_2021}. A rare exception where an AI explanation consisting of an extracted span of text can lead to complementary performance is factoid question answering from trusted sources (illustrated in Figure~\ref{fig:compare} (left)); such a span, even in isolation, does allow verification of answer correctness.

\subsection{A Spectrum of Verification} \label{s:spectrum}
Whether an explanation affords verification of a recommendation is usually not an all or nothing proposition, but rather a point along a spectrum. Characteristics of the human decision maker, the decision making task, and the AI explanation may influence where an explanation lies along this spectrum --- hence, determining its utility. On one extreme are contexts that afford no verification whatsoever, such as problems where the actual outcomes are independent from the problem features, or where all errors made by the AI assistance are random rather than systematic. The other end of the spectrum includes intractable but easily verifiable problems (e.g., the maze solving task, NP-complete decision problems). But even the maze task affords only one-sided verifiability: an explanation depicting an illegal path doesn't imply that no solution exists.

Explanations do not need to provide complete verification of an AI recommendation. Rather, by sufficiently lowering the cost of verification, explanations can provide enough utility to improve team performance~\cite{vasconcelos_when_2023}. Consider a variant of the maze solving task where the AI provides an explanation which highlights 90\% of the path to the exit rather than the complete path. We argue this explanation may reduce the human's cognitive cost of verifying the AI enough to yield complementary performance.

\begin{figure*}[t]
    \centering
    \includegraphics[width=\textwidth]{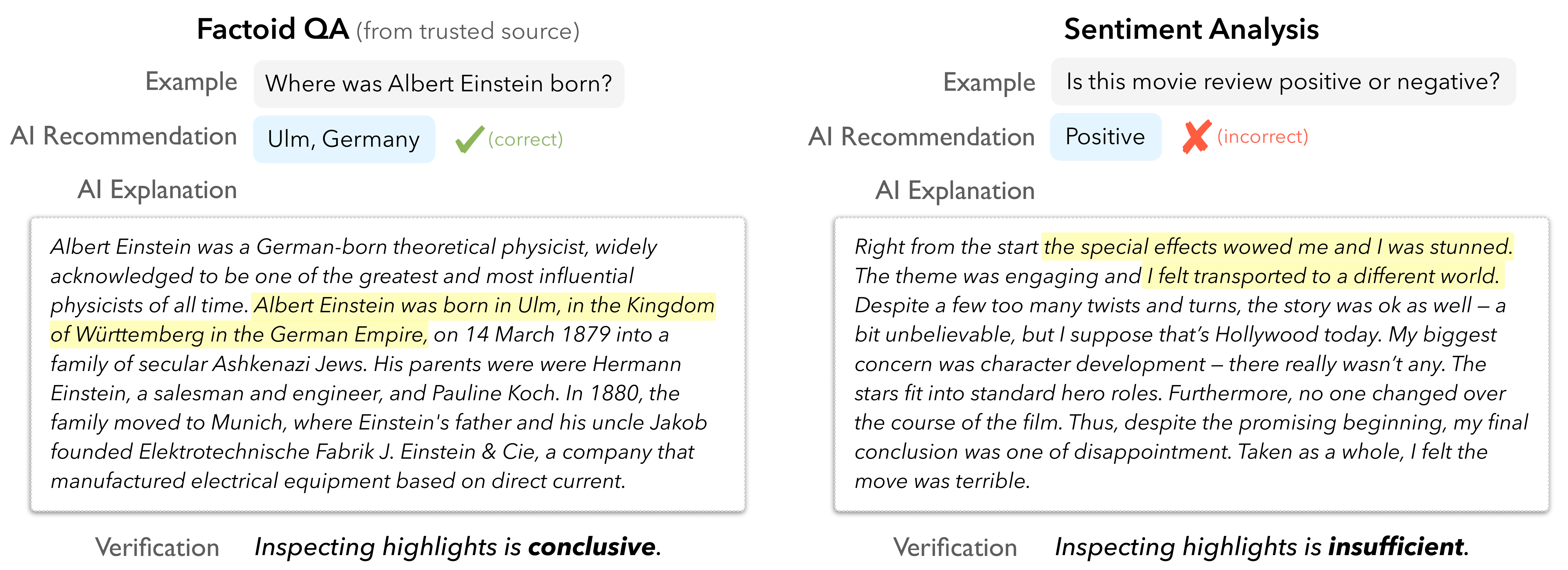}
    \caption{The utility of an explanation is not a simple function of explanation type, as these two examples illustrate. Both use an explanation style that highlight key spans of text, but their ability to induce complementary performance is very different. While these explanations can be useful for factoid QA from trusted sources, they are less successful for sentiment analysis~\protect{\cite{bansal_doesthewhole_2021}}. The difference lies in whether the human can reliably verify the AI's answer by solely inspecting the highlighted text.}
    \Description{}
    \label{fig:compare}
\end{figure*}

Another variation of the maze problem likely leads to a different result. Suppose the question is not just path existence, but optimality. Specifically, consider the following variation on the maze task: given a maze with a start and three alternative exits, which of the exits can be reached with the shortest valid path from the start? The AI recommends a single exit and explains by drawing a path from the start to its specified exit. We conjecture the following hypotheses: 1) this explanation will not lead to complementary performance for this task, since it doesn't significantly reduce the cost of verifying the AI's recommendation. However, 2) if the explanation consisted of the shortest valid paths to \textit{all three} exits, then this \textit{would} yield complementary performance.

With feature-based explanations on real-world decision-making tasks, positive verification is rarely possible. Instead, explanations may be better suited to convey a fatal flaw in the AI's decision making process. For example, if an AI trained to classify lung cancer from CT images justifies its diagnosis by highlighting an artifact outside the lung, that should raise concern~\cite{caruana_intelligible_2015}. Some studies have shown that explanations \textit{can} reveal model error to yield complementary decision making performance (in particular, when the revealed errors are sufficiently egregious to indicate strong signals of model unreliability)~\cite{taesiri_visual_2022, chen_understanding_2023}. However, across most decision making contexts, we believe there are two reasons why such explanations frequently fail to produce complementary performance:
\begin{enumerate}
    \item \textbf{Plausible but wrong:}
    The AI may make a bad decision for a credible reason (e.g., Figure~\ref{fig:compare} right). Hence, explanations which clearly reveal model error to the human are rare.
    \item \textbf{Right for the wrong reason:}
    The AI's recommendation may still be right, even when its explanation is wrong. In this case, the human may incorrectly reject the correct answer. However, even if the human escapes that pitfall, they still need to solve the problem from scratch, and no time is saved.
\end{enumerate}

Many real-world decision making tasks, such as those involving inferences over human action, can further exhibit aleatoric uncertainty, arising from the inherent stochastic nature of the dependency between the observed instance features and the actual decision outcome. In these cases, a correct decision is not entirely determined by features evident to the human or the AI model, but is also influenced by irreducible uncertainty within latent factors (e.g., environmental, social, or societal). For example, one might be able to correlate a criminal's probability for recidivism with factors such as age or gender, but there is no perfect way to predict the future. As a result, the utility of verifying a recommendation becomes limited by the extent of stochasticity present in a task.

In summary, the more easily an explanation enables verification of prediction correctness, the more likely it is to produce complementary performance. Note that both the semantic content of the explanation \textit{and its presentation} (i.e., the user interface which presents the explanation) affect the ease of human verification. However, other factors may also contribute, such as a decision maker's desire to engage with explanations within cognitively effortful tasks (i.e., Need For Cognition~\cite{bucinca_trustorthink_2021}) and the task's inherent cognitive load.

\subsection{Verification and Human Intuitions}
Human verification of an AI recommendation (with or without an explanation) is only possible if \edit{a} decision maker has sufficient task-specific knowledge to characterize solutions. For instance, verification of an explanation in the maze solving task requires implicit assumptions about the human's understanding of the task (e.g., what constitutes a valid path through a maze).

\citet{chen_machineexpl_2023} present a similar framework to ours, articulating how human intuitions can interact with AI explanations to improve decision making. They describe various examples of task-specific human intuitions; for instance, in prostate cancer diagnosis, intuitions might refer to knowledge of the location of the prostate in a medical image and the association between darkness and tumor. In sentiment analysis, intuitions might refer to an understanding of language and its influence on polarity. Through their theoretical framework, they suggest human intuitions can lead to complementary performance in one of two ways: revealing signs of model error and supporting the discovery of novel knowledge.

In line with this theory, the belief that AI explanations could sufficiently interact with task-specific human intuitions to reveal model errors and thus enable complementary performance was a key motivation for providing AI explanations in AI-advised decision making. In practice, however, this type of explanation does not appear to produce complementary performance on most tasks.
\section{What is Appropriate Reliance?}
\label{s:ar}


\begin{figure*}[t]
    \centering
    \includegraphics[width=0.9\textwidth]{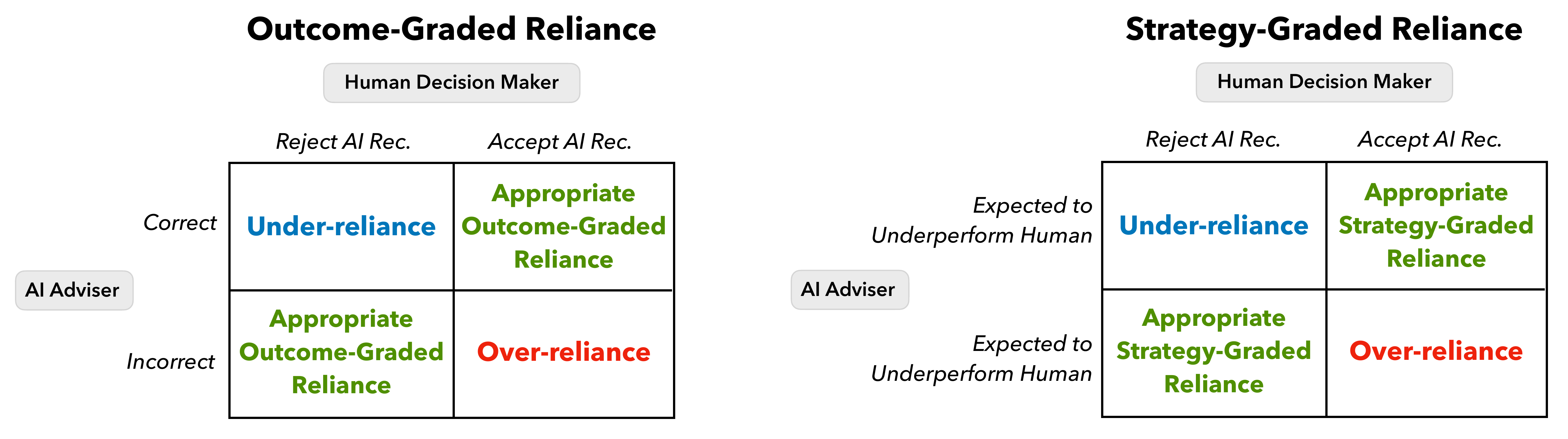}
    \caption{We propose a clarification of two notions of reliance behavior commonly conflated in the literature. 
        \textit{Outcome-graded reliance} is appropriate if the human decision maker accepts an AI recommendation when it is correct and rejects it otherwise; note that this definition is conditioned on the post-hoc correctness of the AI.
        We argue this definition is problematic given its outcome-dependent and nondeterministic nature. In contrast, \textit{strategy-graded reliance} is behavior that defers to the AI when its {\em expected} performance is superior to the human's expected performance.}
    \Description{}
    \label{fig:appropriate_reliance}
\end{figure*}
Previous sections of this paper focus on complementary performance --- an ideal team that performs better than the human or AI alone.  However, many researchers strive instead to design XAI systems that induce appropriate reliance.

Unfortunately, ``appropriate reliance'' is not well-defined within the XAI literature. One common characterization suggests reliance is \textit{appropriate} if the human accepts an AI recommendation when the AI is correct, and rejects the recommendation when the AI is incorrect~\cite{vasconcelos_when_2023, bansal_doesthewhole_2021, bucinca_trustorthink_2021, wang_are_2021, yang_how_2020, schemmer_appropriate_2023}. Reliance is therefore inappropriate when the human accepts the recommendation when the AI is incorrect (over-reliance), and when the human rejects the recommendation when the AI is correct (under-reliance). This characterization captures a notion of \textit{outcome-graded reliance} (see Figure~\ref{fig:appropriate_reliance} left), and is evaluable given actual decision outcomes. Unfortunately, we suggest outcome-graded reliance is inadequate to measure the perceived improvements in decision making performance. There are two major problems with the outcome-graded definition:
\begin{itemize}
    \item \textbf{Post-hoc}: It is impossible to know if one's reliance is `appropriate' until after seeing the final result. Was it `inappropriate' to see the best doctor in the world, if one happens to get an unlucky outcome?
    \item \textbf{Nondeterministic}: Consider two identical examples (e.g., bank loan applications) whose input features are exactly the same, and suppose the AI provides the same recommendation in both cases (e.g., unlikely to default). But suppose that one borrower ends up defaulting, while the other repays the loan. The definition says it is `appropriate' to rely on the AI's advice on one case but not on the other --- even though they are indistinguishable!
\end{itemize}

Instead, consider an alternative definition, \textit{strategy-graded reliance}, where reliance is appropriate if the human accepts an AI recommendation when the AI is {\em expected} to outperform the human, and rejects otherwise (see Figure~\ref{fig:appropriate_reliance} right). Unlike outcome-graded reliance, strategy-graded reliance is neither post-hoc nor nondeterministic; it considers the appropriateness of reliance given the expected relative performance of the human and the AI. The optimal strategy is to rely on the party \textit{most likely} to have the correct answer. A key question here is ``Upon what information is that expectation computed?'' There are several possibilities.

\begin{itemize}
    \item {\bf Past performance:} If past experience shows the AI is more likely to be correct than the human, it might be appropriate to defer to the AI even without information about this particular decision instance. Note this policy cannot produce complementary performance.
    \item Previous characteristics {\bf  + instance features:} Conditioning on the current instance (i.e., specific details of the task at hand) can lead to complementary performance. 
    \edit{For instance, a camera’s automatic exposure mode is a convenient form of automation that is usually accurate. Yet it can err in unusual situations, such as snowy landscapes, so experienced photographers often prefer to override the automation when they recognize these scenarios.}
    When automated, this type of conditioning resembles a human-AI delegation workflow.
    \item Previous characteristics {\bf  + the AI's recommendation:} Conditioning on the AI's recommendation allows the human to adopt a policy of the form ``I know the AI is conservative and very unlikely to err with a false positive, so I will accept positive recommendations and only scrutinize instances when the AI offers a negative recommendation.''
    \item Previous characteristics {\bf  + the AI's explanation:} In this paper, we have argued this condition rarely improves upon the previous strategy, and only when the explanation supports verification.
\end{itemize}

In contrast to complementary performance, which refers to the \textit{team's} measured performance, both notions of reliance define an attribute of the \textit{human's behavior} relative to the AI. Furthermore, while a policy of strategy-guided reliance will {\em hopefully} lead to complementary performance, it's not guaranteed to do so. In particular, if the human's estimates of their relative accuracy (compared to the AI) are poor, team performance may drop.

We believe the strategy-graded definition of reliance is the better objective. To illustrate the shortcomings of outcome-graded reliance, consider a decision making task in which the human is historically 60\% accurate, while the AI is 99.999\% accurate. On any given instance of the task, if the human is uncertain of the answer, is it appropriate to rely on the AI's recommendation? Intuitively, the answer seems a clear `yes'. But if the AI is later found incorrect, the outcome-graded definition says ``Inappropriate,'' while the strategy-graded definition matches intuition and says ``Appropriate.''

Outcome-graded reliance is similar to complementary performance in that both qualities can only be measured post hoc. However, there are subtle differences between these notions, beyond the fact that one measures a pattern of human behavior and the other the performance of a human-AI team. For example, imagine a AI-assisted image classification task with 1\edit{,}000 potential classes. Suppose two individuals, Avery and Blake, are 80\% and 50\% accurate at the task alone, respectively, and the AI is 10\% accurate. Luckily, the AI outputs verifiable explanations, so both Avery and Blake can perfectly tell when the AI is correct. Both follow a policy of accepting the AI's recommendation when it is correct, and otherwise solving the problem themselves. Both their strategies lead to very different expected team performance: 82\% for Avery and 55\% for Blake.\footnote{Expected accuracy can be calculated as $E[acc_{Avery}]$ = $0.1 + 0.8 \times (1 - 0.1) = 0.82$ for Avery and $E[acc_{Blake}]$ = $0.1 + 0.5 \times (1 - 0.1) = 0.55$ for Blake.}
\edit{Yet the definition suggests} both have (near) perfect outcome-graded reliance because their policy dictates following the AI when it is correct and not when it is incorrect. \edit{Moreover, in most scenarios in which the human and AI both err, the reliance is technically `appropriate' as the human is likely to reject the AI recommendation and select a different, albeit incorrect answer.}\footnote{
\edit{There is a small, non-zero probability of each individual trivially selecting the same incorrect answer as the AI. With many possible answers (e.g., 1,000 classes) and assuming that errors are uniformly distributed between the 999 possible incorrect answers, the probability of the human and AI both erring and then selecting the same incorrect answer approaches zero as the answer space grows (e.g., Avery: $((1 - 0.1) \times \frac{1}{999}) \times ((1 - 0.8) \times \frac{1}{999}) < 1e^{-6}$).}
}

Given the limitations of the outcome-graded definition of appropriate reliance, we suggest researchers focus on the strategy-graded, or eschew the term `appropriate reliance' altogether. We argue overall performance is a better objective when evaluating a human-AI team architecture --- whether it involves explanations or not. Furthermore, complementary performance is the holy grail.
\section{Discussion}
\label{s:discuss}

\edit{We have argued that verifiability of AI explanations is crucial for realizing complementary performance in AI-advised decision making. In this section, we consider the connection between verifiability and faithfulness}
(\S\ref{s:faith}), alternative paradigms of human-AI collaboration which might yield complementary performance (\S\ref{s:alt_paradigms}), and the aims of interpretability \edit{\it beyond} complementary performance and decision making support (\S\ref{s:beyond_aiadm}). Finally, we contextualize the verifiability of AI explanations for human-AI decision making among established research on demonstrability and social combination processes in collective group (human-human) decision making (\S\ref{s:group-dm}).

\subsection{Faithfulness \textit{vs.} Verifiability} \label{s:faith}
Some may argue that faithfulness is an intrinsic requirement of explanations. If an explanation does not accurately reflect the underlying prediction or decision making process, it feels deceptive. Certainly, faithful explanations may help a decision maker build a better model of an AI's capabilities over time. And of course, faithfulness is often a desirable property for explanations. We argue, however, that for the context of a single isolated decision, faithfulness is not as important as verifiability.

Consider an unfaithful explanation that facilitates verification. This is hard, but not impossible. Indeed, ~\citet{kahneman_thinking_2011} argues people often solve problems using fast ``System 1'' reasoning and then justify their answers with slower ``System 2'' reasoning. Since the System 2 rationalization fails to match how the answer was discovered, one might call it unfaithful. But if it affords verification then it will suffice to aid the decision maker. On the other hand, consider faithful explanations that do not facilitate verification. Some may be too costly to verify. For instance, showing the numerical weights of a large linear model may provide insight into how a decision was formed, but is infeasible to use for absolute verification. Other explanations may be faithful to the underlying model but fail to capture the entire picture (e.g., Figure~\ref{fig:compare} (right)), or are stymied by uncertainty in the decision making task (e.g., \S\ref{s:spectrum}).

\subsection{Paradigms of XAI-Advised Decision Making} \label{s:alt_paradigms}

\edit{A typical paradigm of AI-advised decision making, and one that we focus our analysis on, involves an AI that provides a recommendation and possibly an explanation.
However, other paradigms are also possible.
For example, 
in} lieu of recommending any singular decision, an {\em evaluative XAI} might instead generate and present evidence to support or refute \edit{several of the options, forcing the human to remain active in forming the final decision.} Such hypothesis-driven support may help mitigate fixation, \edit{anchoring} and associated \edit{automation}  biases by assisting decision makers \edit{to better consider the} trade-offs between multiple options.


\edit{Evaluative XAI is an example of \textit{cognitive forcing}, a class of strategies that has} been \edit{proposed} to improve the efficacy of AI explanations for decision making by introducing interventions which elicit greater decision maker engagement with AI explanations and facilitate deeper thinking about individual task instances~\cite{bucinca_trustorthink_2021, miller_XAIisdead_2023, gajos_do_2022}.
\edit{One associated drawback of cognitive forcing is the increased cognitive load on the decision maker, increasing accuracy at the expense of decision making speed, human effort, or human satisfaction. By reducing the costs of engagement, verifiable explanations may improve the effectiveness of these strategies.}

\edit{Achieving verifiability in practical applications may also involve moving beyond the conventional single-turn, static AI explanations, and toward} interactive explanations that can be generated on-demand and contextualized to individual decision makers. Perspectives from social science theory on explanations tend to support interactive explanations~\cite{miller_explanation_2019, lombrozo_structure_2006, mittelstadt_explaining_2019}, and XAI researchers have similarly encouraged interactivity in AI explanations~\cite{bansal_doesthewhole_2021, adadi_peeking_2018}.
\edit{However,} interactive explanations come at a cost---they \edit{often} require more time and effort to use effectively~\cite{cheng_explaining_2019}, and as a result can further reinforce \edit{detrimental} cognitive biases~\cite{liu_understanding_2021}.

\subsection{Beyond Complementary Performance} \label{s:beyond_aiadm}
\edit{While complementary performance may represent a holy grail for human-AI teams, explanations can also serve other desiderata.} In high-risk settings or where regulatory compliance mandates the presence of human judgment, and yet the AI outperforms humans, it may be desirable for the human-AI team to achieve performance comparable to the AI alone; explanations can help in this regard, since they tend to make the audience trust the recommendation~\cite{bansal_doesthewhole_2021}. Moreover, in some scenarios, objective measures of performance such as accuracy may be of second importance to task efficiency, cognitive load, or decision fairness and transparency.

Rather than supporting decision making over individual instances, AI explanations may be better equipped at facilitating holistic AI processes such as model development and debugging, regulatory audit, and knowledge discovery (Figure \ref{fig:xai_contexts}). To improve models throughout development and after release, explanations have been used to uncover spurious patterns in learned models~\cite{caruana_intelligible_2015} and enable interactive debugging processes~\cite{hildif_zylberajch_2021}. Explanations also offer intelligibility of AI models desirable for ensuring compliance with organizational policies, legal imperatives, and government regulation. For instance, one provision in the European Union's General Data Protection Regulation (GDPR) describes an individual's ``right to explanation,'' particularly in automated decision-making or profiling scenarios that may significantly influence the individual, e.g., legally or financially. Finally, explanations may help decision makers develop new knowledge about a decision making task~\cite{das_leveraging_2020, chen_machineexpl_2023}. If an AI model significantly outperforms human decision makers, and its explanations can effectively convey learned signals or patterns in the task features, then humans may be able to replicate such behavior to improve their own decision making on future instances.

\subsection{Verifiability in Collective Decision Making} \label{s:group-dm}
\edit{AI-advised decision making may be beneficially viewed as an extension of existing social and cognitive psychology research, which has long studied the decision-making processes of human groups.} Many of these \edit{prior} studies together corroborate a generalized theory that groups collectively perform better than the average (and many times, best) individual on many types of problem-solving tasks~\cite{laughlin_1986_demonstrability, laughlin_1980_social, hill_1982_group, laughlin_1966_group, laughlin_2006_groups, carey_2012_groups, stasser_2001_collective, group_1972_steiner, hackman_1975_group, davis_1992_some}.
One relevant line of work postulates a relationship between task type and how \edit{human groups} would arrive at a collective decision~\cite{laughlin_1986_demonstrability, laughlin_1980_social, hill_1982_group}. These studies suggest that the relative superiority of group decision making performance over independent individuals can be explained by the \textit{demonstrability} of a solution for any given task. Initially coined by~\citet{laughlin_1980_social}, demonstrability requires four conditions:
\textit{(1)} Group consensus on a verbal (e.g., vocabulary, syntax) or mathematical system (e.g., primitives, axioms, permissible operations like algebra);
\textit{(2)} Sufficient information for a solution within the system;
\textit{(3)} Group members who cannot solve the problem must have sufficient knowledge to recognize and accept a proposed solution;
\textit{(4)} The correct team member must have sufficient ability, motivation, and time to demonstrate the correct solution to other members.
    
Demonstrability is therefore contingent on the information system and the extent to which the task type affords recognition of correct solutions. \citet{laughlin_1980_social} define a continuum of tasks that exhibit greater or lesser demonstrability, anchored by \textit{intellective} tasks (those with a demonstrably correct answer with a specific verbal or mathematical conceptual system, \edit{similar to tasks with verifiable explanations}) and judgmental tasks (those involving evaluative, behavioral, or aesthetic judgments for which there does not exist a demonstrably correct\edit{, or verifiable,} answer). The criterion for success is different between the tasks: agreement about the correct answer in the former, and achievement of consensus on a collective decision for the latter.
    
\citet{laughlin_1986_demonstrability} showed that human groups, which were working to solve intellective tasks, only required a single group member to have the correct answer, as long as the solution was demonstrable (could be explained in a verifiable manner). Examples of these tasks include puzzles (e.g., ``cannibals and missionaries,'' Tower of Hanoi, or Sudoku) or mathematical problems. Note that these domains are very similar to the cases we addressed earlier (e.g., maze solving, puzzles, or factual QA), which are also intellective and where AI explanations have been shown to yield complementary team performance. In contrast, tasks such as recidivism prediction and loan application evaluation, are examples of judgmental tasks, where no demonstrable (\edit{i.e., verifiable}) solution may exist.


Despite their parallels, current human-AI teams also differ from human groups in classical collective intelligence:

\paragraph{\textbf{Unequal distribution of responsibility and power}}
\edit{In human-AI teams, humans typically retain agency and are responsible for making the final decision. 
Moreover, in contrast to the social psychology literature, which often studies collective reasoning processes with larger groups of humans, the human-AI teams considered by the XAI literature typically consist of just one human and one AI member. 
Future XAI work might consider if there are parallel results to psychologists' observations that the number of human group members required to arrive at a quality collective response is inversely proportional to the demonstrability afforded by the intellective tasks~\cite{laughlin_1986_demonstrability}.} 

\paragraph{\textbf{Rich interaction modalities benefit collective intelligence}}
\edit{Numerous psychology studies have shown that human explanations are interactive, a type of dialog~\cite{miller_explanation_2019}. Yet most existing XAI systems are static and single-turn.
Moreover, current interaction paradigms are unilateral; explanations are provided by AI models to the human, but rarely are AI systems able to incorporate or adjust to a human explanation. 
Recent interactive explanations have begun to revise this narrative~\cite{Weld2018TheCO}, with abundant opportunities for future work.}

\section{Conclusion} \label{s:conclude}

There are many reasons to develop AI systems that can explain themselves, but it has become apparent that different uses require different types of explanations. It is therefore important to ensure a given type of explanation is well-aligned with its use. In this paper, we focused specifically on the utility of AI explanations for instance-level human-AI decision making. We proposed that explanations can engender complementary performance in decision making to the extent the explanation allows a human decision maker to verify correctness of the AI's recommendation. Unfortunately, most kinds of explanations---notably those which elucidate the AI's reasoning process---\textit{do not} facilitate verification and hence are not helpful for decision making. We further teased apart two notions of appropriate reliance, outcome-graded and strategy-graded, advocating use of the strategy-graded formulation, which sidesteps the problem of nondeterminism and aligns with intuitions of appropriateness. In the pursuit of complementary team performance, we should reevaluate the extent to which the AI explanations we often integrate into our decision making systems facilitate verification.

\begin{acks}
    This research was supported by ONR grant N00014-21-1-2707 and the Allen Institute for Artificial Intelligence (AI2).
We thank Gagan Bansal, Tongshuang Wu, and the members of the Lab for Human-AI Interaction at the University of Washington for formative discussions.
Chacha Chen,
Hal Daumé III,
Madeleine Grunde-McLaughin, 
\edit{Rao Kambhampati,}
Sunnie Kim,
Raymond Mooney,
Giang Nguyen,
Marissa Radensky, 
Mark Riedl,
Suzanna Sia,
\edit{and Chenhao Tan}
provided helpful comments \edit{on the manuscript}.
We also thank the anonymous reviewers for their insightful feedback.
\end{acks}

\bibliographystyle{ACM-Reference-Format}
\bibliography{main}

\end{document}